\documentclass[wcp]{jmlr}

\usepackage{longtable}

\usepackage{booktabs}

\usepackage{lineno}
\usepackage{tikz}
\usetikzlibrary{arrows.meta}
\usepackage[all]{hypcap}
\setcitestyle{citesep={,}} 
\makeatletter
\let\Ginclude@graphics\@org@Ginclude@graphics 
\makeatother

\jmlryear{2026}
\jmlrworkshop{ACML 2026}

\title[Big Models, Small Gains]
  {Big Models, Small Gains:\\
   Frozen Forecasters as Classifiers on MONSTER}

 \author{\Name{Rishit Jain} \Email{rishit2006@icloud.com}\\
  \Name{Rishit Kar} \Email{emailrishitkar@gmail.com}\\
  \Name{Chetashri Bhadane} \Email{chetashri.bhadane@djsce.ac.in}\\
  \Name{Nilesh Patil} \Email{nilesh.p@djsce.ac.in}\\
  \addr Department of Computer Engineering, Dwarkadas J. Sanghvi College of Engineering,\\ University of Mumbai, India}

\editors{}

\begin{document}

\makeatletter
\let \@jmlrpages \@empty
\makeatother

\maketitle

\begin{abstract}
 Forecasting foundation models are increasingly used as frozen feature extractors for time series classification, but it remains unclear whether their pretrained representations offer an advantage over strong classifiers that do not require pretraining. This raises an important question: do these representations provide useful information beyond what the model architecture alone can capture, particularly when labelled data are scarce? We present a controlled benchmarking of two forecasting models, Reverso and Chronos-Bolt, on six large datasets from the MONSTER benchmark, using 1\%, 10\%, and 100\% of the available training labels. We compare their frozen representations against four established time series classification methods and random-weight versions of the same forecasting models. Our results show that pretraining improves classification accuracy over random-weight counterparts on five of the six datasets, with gains of up to 23 percentage points. However, the frozen representations do not reliably outperform the strongest non-pretrained baseline in any setting: they perform worse in 15 of 17 settings and tie in the remaining two. Further experiments show that restoring amplitude information recovers part of the performance gap, while fine-tuning and forecast-error features do not close it. These findings highlight the importance of strong task specific baselines and random weight controls when evaluating forecasting foundation models for classification. Our code is available at \url{https://github.com/Rachet2006/frozen-forecasters-monster}.
\end{abstract}

\section{Introduction}\label{sec:intro}
Foundation models pretrained to forecast time series, such as Chronos
\citep{ansari2024chronos}, TimesFM \citep{das2024timesfm}, Moirai
\citep{woo2024moirai} and Reverso \citep{reverso2026}, are increasingly
reused for classification. The usual approach keeps the model frozen, pools
its hidden states over time (mean or max pooling) and trains a standard
classifier on the resulting embedding \citep{auer2025frozen,reverso2026}.
If pretraining on large collections of unlabelled series yields general
features, they should matter most when labels are scarce.

It is not yet clear whether this holds against strong methods that need no
pretraining, such as MiniRocket and Quant, which are among the most accurate
time series classifiers \citep{middlehurst2024bakeoff,dempster2025monster}.
Existing evaluations have three gaps (Section~\ref{sec:related}). First,
they use the UCR and UEA archives, whose datasets are small, with median
sizes of 217 and 255 series \citep{dempster2025monster}. Second, they
compare mainly with other foundation models. Third, they do not separate the
effect of the pretrained weights from the effect of the network architecture
itself. We address all three. We evaluate two frozen forecasters, Reverso
(2.6M parameters) and Chronos-Bolt (205M), on six large MONSTER datasets
\citep{dempster2025monster} at three label budgets (1\%, 10\% and 100\% of
the training labels). We compare them with four strong baselines and with a
random-weight copy of each forecaster, which isolates the effect of
pretraining (Figure~\ref{fig:design}). Both forecasters are large compared
with the baselines, which learn no weights before seeing the labels and
extract features up to 1{,}000 times faster (Appendix~\ref{apd:cost}).

Pretraining matters, but the frozen models never reliably beat the best
baseline, and are often furthest behind when labels are scarcest. Our contributions are:
\begin{enumerate}\setlength{\itemsep}{1pt}\setlength{\parskip}{0pt}
\item A benchmarking study comparing frozen forecasting foundation models with MiniRocket, MultiRocket, Hydra and Quant on MONSTER across three label budgets.
\item A random-weight control to distinguish the benefits of forecasting pretraining from those provided by the model architecture alone.
\item An analysis of the performance gap through experiments on amplitude statistics, fine-tuning, and the use of forecast errors as additional input features for a baseline classifier.
\item Two evaluation pitfalls: a gain from the architecture that looks like
a gain from pretraining without a random-weight copy, and kernel seed
variation that confidence intervals over folds miss.
\end{enumerate}

\section{Related Work}\label{sec:related}
\paragraph{Forecasting foundation models as feature extractors.}
\citet{auer2025frozen} froze forecasting foundation models, pooled their
hidden states over time and trained a random forest on the embeddings. This
matched dedicated classification foundation models on the UCR and UEA
archives. Reverso's authors followed the same approach, adding max pooling,
patch statistics and an embedding of the differenced series
\citep{reverso2026}. Neither study compares with strong task-specific
classifiers such as ROCKET and its successors, an omission that
\citet{auer2025frozen} note themselves.

\paragraph{Strong baselines without pretraining.}
ROCKET applies thousands of random convolutional kernels to a series and
pools each output into features for a linear classifier
\citep{dempster2020rocket}. MiniRocket makes this almost deterministic and
much faster \citep{dempster2021minirocket}. MultiRocket also applies the
kernels to the first difference of the series and pools each output in four
ways instead of one \citep{tan2022multirocket}. Hydra groups the kernels and
counts how often each kernel is the best match within its group
\citep{dempster2023hydra}. Quant takes quantiles of the series over fixed
intervals and classifies them with extra trees \citep{dempster2024quant}.
None of these methods needs pretraining, and all rank among the strongest on
standard benchmarks \citep{middlehurst2024bakeoff,dempster2025monster}. Where
foundation models have been compared with them, the specialised methods match
or beat the foundation models \citep{isiacik2025empirical,xu2025specialized}.
When both feed the same TabPFN classifier \citep{hollmann2025tabpfn}, ROCKET
features also beat embeddings from the classification foundation models
Mantis \citep{feofanov2025mantis} and MOMENT \citep{goswami2024moment} on 103
UCR datasets \citep{rocketpfn2026}. 

\paragraph{Random-weight controls.}
A network with random weights can already be a good feature extractor
\citep{jarrett2009best,saxe2011random}, as ROCKET's random kernels show for
time series, and random encoders have been proposed as baselines for
sentence representations \citep{wieting2019notraining}. In forecasting,
\citet{tan2024llmuseful} found that randomly re-initialising the language
model inside LLM-based forecasters left their accuracy largely unchanged. To
our knowledge, no study has applied this control to forecasting foundation
models used for classification.

\section{Methodology}\label{sec:method}
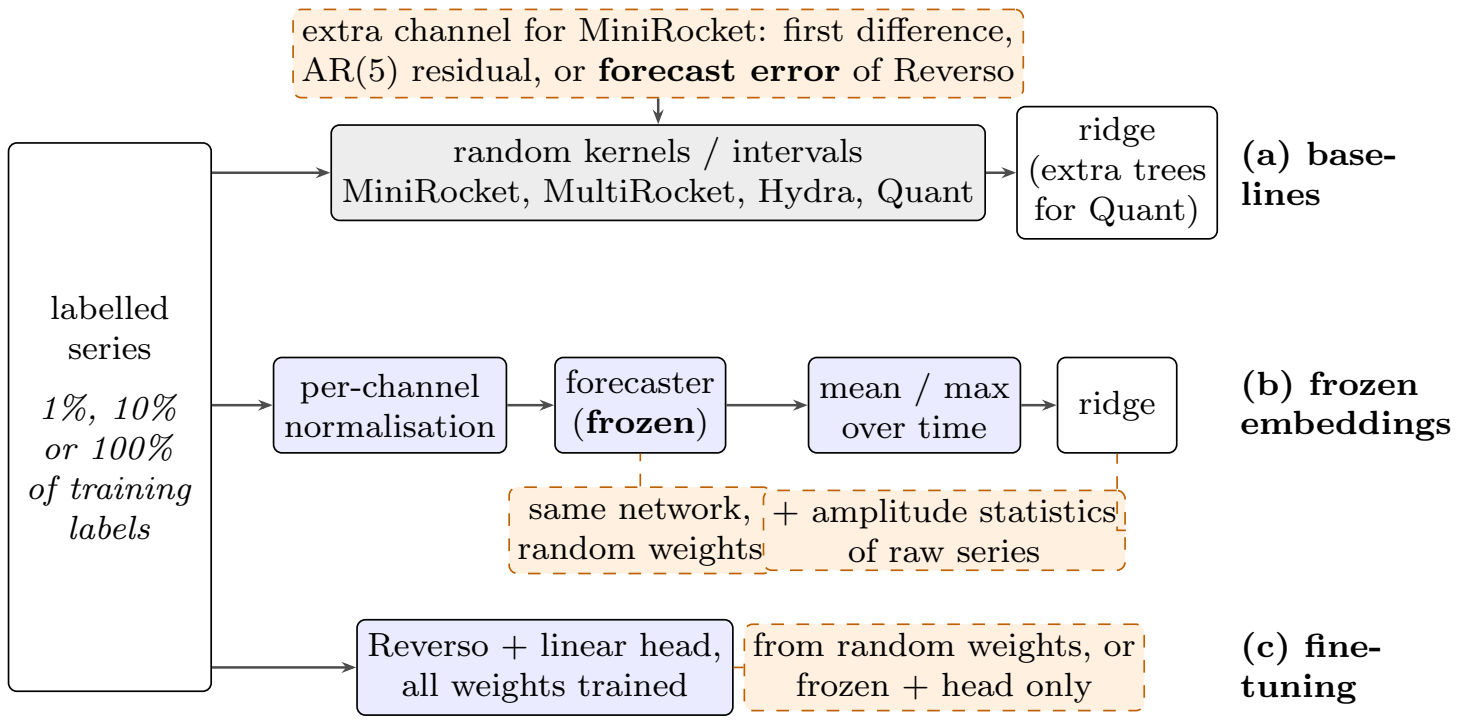
\begin{figure}[t]
\centering
\resizebox{0.64\textwidth}{!}{
\begin{tikzpicture}[
  font=\scriptsize,
  >={Stealth[length=3.5pt]},
  box/.style={draw, rounded corners=2pt, align=center, inner sep=2.5pt, minimum height=8mm},
  base/.style={box, fill=black!7},
  fm/.style={box, fill=blue!8},
  head/.style={box, fill=white, minimum width=10mm},
  ctl/.style={draw=orange!75!black, dashed, rounded corners=2pt, fill=orange!12,
              align=center, inner sep=2pt, minimum height=5.5mm},
  lab/.style={font=\scriptsize\bfseries, anchor=west, align=left},
  arr/.style={->, semithick, black!70},
]
\node[box, minimum height=46mm, minimum width=17mm] (x) at (0,-0.1)
  {labelled\\series\\[5pt]\textit{1\%, 10\%}\\\textit{or 100\%}\\\textit{of training}\\\textit{labels}};

\node[base, minimum width=42mm] (rk) at (4.6,1.95)
  {random kernels / intervals\\MiniRocket, MultiRocket, Hydra, Quant};
\node[ctl, minimum width=42mm] (ch) at (4.6,2.95)
  {extra channel for MiniRocket: first difference,\\AR(5) residual, or \textbf{forecast error} of Reverso};
\node[head] (h1) at (8.45,1.95) {ridge\\(extra trees\\for Quant)};
\draw[arr] (x.east |- rk) -- (rk);
\draw[arr] (ch) -- (rk);
\draw[arr] (rk) -- (h1);
\node[lab] at (9.35,1.95) {(a) base-\\lines};

\node[fm] (nrm) at (2.35,0) {per-channel\\normalisation};
\node[fm] (net) at (4.45,0) {forecaster\\(\textbf{frozen})};
\node[fm] (pool) at (6.75,0) {mean / max\\over time};
\node[head] (h2) at (8.45,0) {ridge};
\node[ctl] (twin) at (4.45,-1.05) {same network,\\random weights};
\node[ctl] (amp) at (7.0,-1.05) {+ amplitude statistics\\of raw series};
\draw[arr] (x.east |- nrm) -- (nrm);
\draw[arr] (nrm) -- (net);
\draw[arr] (net) -- (pool);
\draw[arr] (pool) -- (h2);
\draw[dashed, orange!75!black] (twin) -- (net);
\draw[dashed, orange!75!black] (amp) -| (h2);
\node[lab] at (9.35,0) {(b) frozen\\embeddings};

\node[fm, minimum width=31mm] (ft) at (3.65,-2.2) {Reverso + linear head,\\all weights trained};
\node[ctl, minimum width=24mm] (sc) at (7.0,-2.2) {from random weights, or\\frozen + head only};
\draw[arr] (x.east |- ft) -- (ft);
\draw[dashed, orange!75!black] (ft) -- (sc);
\node[lab] at (9.35,-2.2) {(c) fine-\\tuning};
\end{tikzpicture}}
\caption{Study design. (a) Baselines without pretraining. (b) Frozen forecaster
embeddings. (c) Reverso fine-tuned end to end. Dashed boxes: controls and
variants.}
\label{fig:design}
\end{figure}

All three pipelines (Figure~\ref{fig:design}) use the same labelled subsets
and test folds (details in Appendix~\ref{apd:impl}).

\paragraph{Baselines without pretraining.}
We use the aeon implementations \citep{middlehurst2024aeon} of MiniRocket,
MultiRocket, Hydra and Quant. The first three are fitted unsupervised on each
fold's training series, and their features feed a ridge classifier
\citep{hoerl1970ridge,pedregosa2011sklearn} whose regularisation strength is
chosen by cross-validation. Quant uses its own extra-trees classifier
\citep{geurts2006extra}.

\paragraph{Frozen forecaster embeddings.}
We use two forecasters that differ in almost every respect. Reverso
\citep[2.6M parameters,][]{reverso2026} reads one time step per token, which
suits our short series, and its authors already evaluate it for
classification. Chronos-Bolt \citep[base, 205M
parameters,][]{ansari2024chronos,chronosbolt2024} is a transformer about 80
times larger that reads patches of 16 steps. Each channel of a series is
passed separately through the frozen model. For Reverso we pool the hidden
states of all eight layers, following its authors' recipe, and for
Chronos-Bolt those of the final encoder layer, the usual way to embed with
it. We use mean or max pooling over time, concatenate the pooled features
across channels and train the same ridge classifier as for the baselines, so
that the comparison differs only in the features. A random forest, as used by
\citet{auer2025frozen}, did no better (Appendix~\ref{apd:tables}).

Both models normalise each input channel, Reverso by scaling it to $[0,1]$
and Chronos-Bolt by standardising it \citep{kim2022revin}. This removes the
level and scale of the series, which can carry information useful for
classification \citep{liu2022nonstationary}. Following Reverso's recipe
\citep{reverso2026}, we therefore also append amplitude statistics $S$: the
mean, standard deviation, minimum and maximum of each of eight equal
segments of each raw channel.

\paragraph{Random-weight copies.}
For each forecaster we build a single copy with the same architecture and
input normalisation but randomly initialised weights (seed 0), and extract
features in exactly the same way. The accuracy difference between the
pretrained model and its copy measures what pretraining contributes.

\paragraph{Fine-tuning.}
We attach a linear head to
Reverso's last layer and fine-tune all weights with a recipe fixed in
advance. Two controls use the same recipe: training the same network from
random initialisation, and linear probing, which trains only the head on the
frozen backbone.

\paragraph{Forecast errors as an extra input.}
Finally, we test whether a forecaster can help a baseline indirectly. For
each input channel, MiniRocket receives one extra channel holding Reverso's
forecast error $x_t - \hat{x}_t$, where $\hat{x}_t$ is Reverso's prediction
of $x_t$ from earlier values only. As cheap controls, the extra channel
instead holds the first difference $x_t - x_{t-1}$, following MultiRocket
\citep{tan2022multirocket}, or the residual of a fifth-order autoregressive
model, AR(5), fitted to each series. The AR(5) residual is itself the error
of a simple linear forecaster, so it is the cheap counterpart of Reverso's
forecast error. Its order was fixed in advance and kept small to suit short
series. We fixed this design and the rule for judging it
(Appendix~\ref{apd:impl}) before running it, using UCIActivity, CrowdSourced
and InsectSound as pilot datasets.

\section{Experimental Setup}\label{sec:setup}
We use six MONSTER datasets \citep{dempster2025monster} from five domains:
smartphone motion (UCIActivity, WISDM2), EEG (CrowdSourced), insect wingbeat
audio (InsectSound), vehicle sensors (FordChallenge) and seismograms (LenDB).
They hold 10{,}299 to 626{,}180 series (Table~\ref{tab:data}), and we use
their predefined folds (five per dataset, two for LenDB). In each fold,
five seeds draw stratified subsets of 1\% and 10\% of the training labels,
and the 100\% budget uses all of them. Methods are compared by paired
per-fold differences with 95\% t-intervals. A difference counts as reliable
when its interval excludes zero and as a tie otherwise. The best embedding
is chosen from six variants and the best baseline from four, which favours
the embeddings, so the reported gaps are conservative.
Appendix~\ref{apd:setup} gives full details about the setup.

\section{Results}\label{sec:results}
\begin{figure}[t]
\centering
\includegraphics[width=\textwidth]{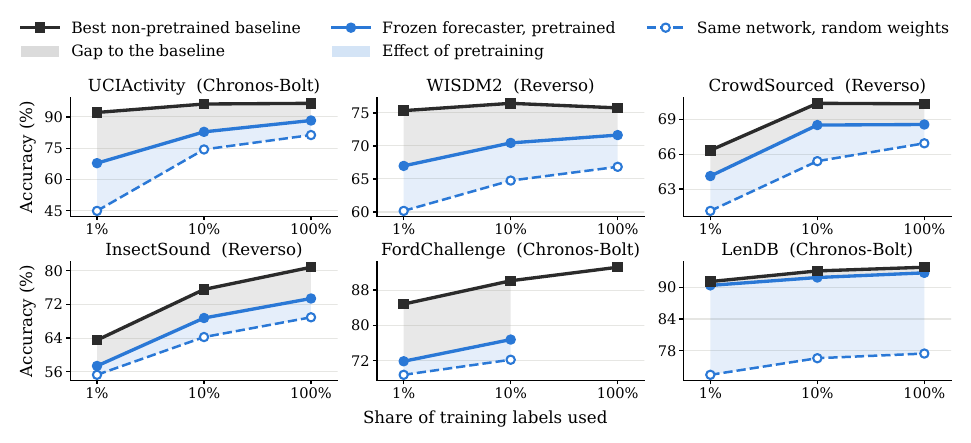}
\caption{Accuracy against the share of training labels (mean over folds,
y-axes differ between panels). Black: best baseline at each budget. Solid
blue: frozen forecaster with mean pooling, showing on each dataset the
stronger of the two models (named in the panel title), the most favourable
case for the foundation models. Dashed blue: the same network with random
weights. The grey band is the gap to the baseline and the blue band the
effect of pretraining. Embeddings were not run at 100\% on FordChallenge.
Results for both models are in Tables~\ref{tab:full1}--\ref{tab:full100}.}
\label{fig:results}
\end{figure}
\begin{table}[t]
\centering\small
\setlength{\tabcolsep}{5pt}
\renewcommand{\arraystretch}{1.15}
\caption{Main results (accuracy points, mean over folds). Gap: best frozen
embedding minus best baseline (negative means the embedding is worse).
Pretraining effect: pretrained minus random-weight embedding, mean pooling
($^{\dagger}$ max pooling). All differences are reliable except those marked
$^{=}$ (tie). -- not run.}
\label{tab:main}
\begin{tabular}{l rrr rrr rrr}
\toprule
 & \multicolumn{3}{c}{Gap to best baseline} & \multicolumn{6}{c}{Pretraining effect} \\
\cmidrule(lr){2-4}\cmidrule(lr){5-10}
 & & & & \multicolumn{3}{c}{Reverso} & \multicolumn{3}{c}{Chronos-Bolt} \\
\cmidrule(lr){5-7}\cmidrule(lr){8-10}
Dataset & 1\% & 10\% & 100\% & 1\% & 10\% & 100\% & 1\% & 10\% & 100\% \\
\midrule
UCIActivity   & $-16.9$ & $-5.1$ & $-1.9$ & $+10.4$ & $+7.5$ & $+6.0$ & $+22.9$ & $+8.5$ & $+7.0$ \\
WISDM2        & $-4.8$ & $-4.1$ & $-2.4$ & $+6.8$ & $+5.7$ & $+4.8$ & $+6.9$ & $+4.1$ & $+3.4$ \\
CrowdSourced  & $+0.3^{=}$ & $-1.3$ & $-0.3^{=}$ & $+3.0$ & $+3.1$ & $+1.6^{=}$ & $+3.9^{=}$ & $+2.0^{=}$ & $-0.5^{=}$ \\
InsectSound   & $-3.9$ & $-6.0$ & $-7.2$ & $+2.1$ & $+4.5$ & $+4.4$ & $+4.8$ & $+6.1$ & $+6.0$ \\
FordChallenge & $-11.1$ & $-8.9$ & -- & $+1.2$ & $+2.3$ & -- & $+3.1$ & $+4.6$ & -- \\
LenDB         & $-0.7$ & $-1.2$ & $-1.1$ & $+10.5^{\dagger}$ & $+10.4^{\dagger}$ & $+10.4^{\dagger}$ & $+16.9$ & $+15.2$ & $+15.2$ \\
\bottomrule
\end{tabular}
\end{table}

\paragraph{Frozen embeddings never reliably beat the best baseline.}
In 15 of the 17 settings the best frozen embedding is reliably worse than
the best baseline, and the other two, both on CrowdSourced, are ties
(Table~\ref{tab:main}, Figure~\ref{fig:results}). Scarce labels do not
favour the foundation models. On UCIActivity, WISDM2 and FordChallenge the
gap is widest at 1\% of the labels and narrows as labels are added, the
opposite of what the hypothesis in Section~\ref{sec:intro} predicts. On
InsectSound the gap instead widens with more labels, and on CrowdSourced and
LenDB it barely changes. The closest result is on LenDB, where Chronos-Bolt
comes within about one point of Quant and beats every random-kernel method.

\paragraph{Pretraining helps compared with random weights.}
On five of the six datasets, each pretrained model beats its random-weight
copy in every fold, by up to 23 points (Chronos-Bolt on UCIActivity at 1\%).
On UCIActivity and WISDM2 this benefit is largest when labels are scarcest,
so pretraining does help most where the hypothesis expects, just not
enough to match the baselines.
CrowdSourced is the exception, where the effect is reliable only for Reverso
at 1\% and 10\%. The random-weight copies are themselves strong. On
InsectSound, Reverso's copy reaches 55 to 69\% accuracy, against 21 to 26\%
for the amplitude statistics alone, so most of the pretrained model's
accuracy is available without pretraining.

\paragraph{Lost amplitude explains part of the gap.}
Appending the amplitude statistics $S$ recovers 6 to 9 points on
UCIActivity and 1 to 4 points on WISDM2 and FordChallenge, but changes
accuracy by at most 2 points on the other three datasets
(Tables~\ref{tab:full1}--\ref{tab:full100}). Adding $S$ to MiniRocket, which
already sees the raw amplitudes, changes its accuracy by at most 0.3 points
(folds 0--1). The statistics therefore restore what normalisation removes
rather than add new information.

\paragraph{Fine-tuning does not close the gap.}
Fine-tuned Reverso beats the same network trained from random weights in 17
of 18 runs at 1\% labels, so pretraining helps here too, but it stays below
the best baseline in every case, by 1.5 points at the closest (InsectSound,
1\%) (Table~\ref{tab:ft}). At 10\% labels it is even less accurate than the
frozen embedding with $S$ on UCIActivity and CrowdSourced, at about ten times
the compute (Table~\ref{tab:cost}).

\paragraph{Forecast errors are no more useful than a first difference.}
The forecast-error channel has a lower mean accuracy than the
first-difference channel in all 12 settings where it was run, and fails its
pre-specified rule on all three pilot datasets (Table~\ref{tab:fe}). It also
costs 40 minutes to 2.7 hours of GPU time per fold, against seconds for the
first difference (Table~\ref{tab:cost}).

\paragraph{Two evaluation pitfalls.}
First, during exploration, adding MiniRocket features computed on Reverso's
hidden states improved MiniRocket with a first-difference channel. On folds
not used during exploration the gain held only on InsectSound, and Reverso's
random-weight copy gave the same gain ($+2.7$ against $+2.6$ points at 1\%
labels), while as many extra features computed on the raw input gave none.
Without the random-weight control, this gain from the architecture would
have been credited to pretraining. Second, changing only MiniRocket's random
seed moves its accuracy on CrowdSourced at 1\% labels from 59.1 to 65.5\%
(folds 0--1), while on InsectSound and UCIActivity no random-kernel method
moves by more than 1.4 points. Confidence intervals over folds with a single
kernel seed miss this variation.

\section{Conclusion and Limitations}\label{sec:conclusion}
Pretraining improves classification performance: on five of the six datasets, both frozen forecasters outperform their random-weight counterparts in every fold. However, on the larger MONSTER datasets, neither consistently outperforms the simpler baselines at any label budget, and the gap is largest at low label budgets on three datasets. Adding amplitude statistics recovers some of this gap, while fine-tuning and forecast-error inputs provide no consistent further gains. These findings highlight the need to evaluate foundation-model features against strong baselines and random-weight controls. Our study is limited to two forecasters, one random initialisation per model, representations pooled across different layers, accuracy as the sole metric, and fine-tuning experiments on three datasets with two folds. We also use a subsample of LenDB and a smaller-than-default MultiRocket configuration.

\acks{We thank the mentors of DJSACM, especially Mihir Panchal, for their guidance and feedback throughout this work.}

\FloatBarrier
\bibliography{acml26}

\appendix

\section{Experimental Setup Details}\label{apd:setup}
\paragraph{Datasets.}
Table~\ref{tab:data} lists the six datasets: smartphone motion
(UCIActivity, WISDM2), EEG (CrowdSourced), wingbeat audio (InsectSound),
vehicle sensors (FordChallenge) and seismograms (LenDB)
\citep{ford_challenge,weiss2012personalization,anguita2013har,chen2014insects,magrini2020lendb, crowdsourced_eeg}.
MONSTER's predefined folds keep subjects, driving sessions or seismic
networks apart between training and test. LenDB has 1.24M series and two
folds. We keep both test folds whole, draw one stratified pool of
100{,}000 training series from the rest as LenDB's full training set, and
multiply all values by one global constant because Reverso's min--max
scaling clamps very small ranges. Reverso's pretraining corpus is mainly
energy, traffic, weather, sales, web, cloud and epidemiology data, with no
activity, EEG or audio dataset by name.

\begin{table}[!ht]
\centering\footnotesize
\caption{Datasets. Labels at 1\%: size of the smallest 1\% training subset
over all folds and, in brackets, the smallest class count on folds 0--1. LenDB:
our 100{,}000 training series plus both test folds.}
\label{tab:data}
\begin{tabular}{llrrrrr}
\toprule
Dataset & Domain & Series & Channels & Length & Classes & Labels at 1\% \\
\midrule
UCIActivity & motion & 10{,}299 & 9 & 128 & 6 & 78 (10) \\
WISDM2 & motion & 149{,}034 & 3 & 100 & 6 & 946 (22) \\
CrowdSourced & EEG & 12{,}289 & 14 & 256 & 2 & 94 (47) \\
InsectSound & audio & 50{,}000 & 1 & 600 & 10 & 400 (40) \\
FordChallenge & vehicle & 36{,}257 & 30 & 40 & 2 & 290 (105) \\
LenDB (subset) & seismic & 626{,}180 & 3 & 540 & 2 & 1{,}000 (488) \\
\bottomrule
\end{tabular}
\end{table}

\FloatBarrier
\paragraph{Label budgets.}
For each fold and each of five seeds we draw stratified subsets with 1\% and
10\% of the training series. The 100\% budget uses the full training set.
Every method sees the same subsets. Even at 1\%, subsets hold 78 to
1{,}220 labelled series depending on dataset and fold, so this is not a
few-shot setting.

\paragraph{Statistics.}
We average the five seeds within a fold, treat each fold as one measurement
and compare methods by paired per-fold differences with 95\% t-intervals. A
difference is reliable when its interval excludes zero and is otherwise a
tie. For LenDB, which has two folds, it is reliable when it has the same sign
in both folds. In each setting, the best baseline is the best
of the four baselines and the best embedding the best of six variants (mean
pooling, max pooling, or mean pooling with $S$, for each model), both chosen
by mean accuracy over folds. We report the gap (best embedding minus best
baseline) and the pretraining effect (pretrained minus random-weight
embedding), where each accuracy is averaged over the seeds of one fold and
the differences are paired by fold.

\paragraph{Coverage.}
The main grid covers all datasets and budgets except the embeddings at
100\% on FordChallenge, too large for our hardware, leaving 17
settings. The forecast-error channel was run on four datasets
and fine-tuning on three, both with two folds.

\FloatBarrier
\section{Implementation Details and Reproducibility}\label{apd:impl}
\paragraph{Baselines.}
MiniRocket uses its default of about 10{,}000 features (9{,}996). MultiRocket uses 1{,}250
kernels instead of the default 6{,}250, which matches MiniRocket's feature
budget of about 10{,}000 and fits in memory on the largest datasets. Hydra
uses its default settings. All three come from aeon 1.6.0 with random seed 0
and are fitted on the fold's unlabelled training series. Features are
standardised on the training subset, and the ridge regularisation strength
is chosen by leave-one-out cross-validation from 13 values spaced
logarithmically between $10^{-3}$ and $10^{5}$. For more than 20{,}000
training series we use an equivalent eigendecomposition-based solver, which
gave identical predictions in a check against scikit-learn. Quant uses its
default intervals (depth 6, quantile divisor 4) and an extra-trees
classifier with 200 trees, a feature fraction of 0.1 and the entropy
criterion.

\paragraph{Extra input channels.}
Forecasts are made in blocks of 16 steps: for $t_0 = 16, 32, \ldots$ Reverso
forecasts the next 16 values from $x_{<t_0}$ only, and the error is set to
zero in the first block. The AR(5) model is fitted to each series and
channel separately. MiniRocket keeps its budget of 9{,}996 features, so an
extra channel replaces some features rather than adding to them. The
decision rule required the forecast-error channel to beat the better cheap
channel with a 95\% interval excluding zero, at 1\% or 10\% labels, on at
least two of the three pilot datasets (UCIActivity, CrowdSourced,
InsectSound).

\paragraph{Embeddings.}
Reverso has eight layers that alternate convolution and DeltaNet attention,
with width 128 and a context of 2{,}048 steps. Series are padded at the
start with their first value, and only the states of the real positions are
pooled, giving 1{,}024 features per channel. For Chronos-Bolt (chronos
2.3.2) we pool the final encoder states over the patch tokens, excluding its
summary token, giving 768 features per channel. Random-weight copies use
each model's own initialisation with seed 0.

\paragraph{Fine-tuning.}
The head takes the mean and maximum over time of Reverso's last layer (256
values per channel), followed by layer normalisation and a zero-initialised
linear layer. We train with AdamW (learning rate $10^{-4}$ for the network
and $10^{-3}$ for the head, weight decay 0.01), a cosine schedule to zero,
gradient clipping at 1.0 and bfloat16 autocast, for 50 epochs at 1\% labels
and 20 at 10\%, with about 64 channel-series per update through gradient
accumulation. There is no validation split, and the model after the last
epoch is evaluated.

\paragraph{Reproducibility.}
Reverso was run from its authors' repository with the pure-PyTorch model,
in which we replaced the step-by-step DeltaNet recurrence with an exact
chunked form (chunks of 64). Layer outputs agree within $1.7\times10^{-6}$
and forecasts within a relative difference of $5\times10^{-4}$. LenDB values
are multiplied by $10^{6}$ for all methods, because Reverso's min--max
scaling clamps ranges below $10^{-5}$. On LenDB, Hydra's inputs are divided by
one further global constant to avoid overflow when storing features, and
Quant's features are stored as $\operatorname{sign}(v)\log(1+|v|)$, a
monotone map that leaves the tree splits unchanged up to rounding. Software:
Python 3.11, PyTorch 2.6, aeon 1.6.0, scikit-learn 1.9.1 and chronos 2.3.2,
on one machine with an 8\,GB GPU and 16\,GB of RAM. Code, per-run results and the scripts that build the tables are available at
\url{https://github.com/Rachet2006/frozen-forecasters-monster}.

\FloatBarrier
\section{Additional Results}\label{apd:tables}
\paragraph{Fine-tuning.}
Table~\ref{tab:ft} gives the fine-tuning runs. Fine-tuning beats training
from scratch in 17 of 18 paired runs at 1\% and in 5 of 6 at 10\%. Linear
probing is weaker than the frozen ridge embedding
with amplitude statistics on UCIActivity (61.3 against 67.3\%) and
InsectSound (42.8 against 56.8\%) at 1\%, although the head sees only the
last layer while the ridge embedding uses all eight.

\begin{table}[!ht]
\centering\footnotesize
\caption{Fine-tuning Reverso (accuracy \%, folds 0--1 only, same label subsets for
all columns, 6 runs at 1\%, 2 at 10\%). Best baseline and $R+S$ are
recomputed on these runs, so they differ from Table~\ref{tab:main}.
Differences in the text are computed before rounding. $R+S$: frozen Reverso,
mean pooling, with amplitude statistics. FT: fine-tuned from pretrained
weights. SC: same recipe from random weights. LP: frozen network with the
same head. FT $>$ SC: runs in which fine-tuning beats training from scratch.}
\label{tab:ft}
\begin{tabular}{llrrrrlr}
\toprule
Dataset & Labels & FT & SC & LP & FT $>$ SC & Best baseline & $R+S$ \\
\midrule
UCIActivity & 1\% & 66.9 & 59.2 & 61.3 & 6/6 & MiniRocket 93.8 & 67.3 \\
 & 10\% & 84.5 & 80.0 & 77.9 & 2/2 & MiniRocket 97.0 & 91.6 \\
CrowdSourced & 1\% & 64.4 & 61.2 & 65.6 & 5/6 & MultiRocket 67.0 & 64.2 \\
 & 10\% & 68.9 & 69.9 & 69.4 & 1/2 & Hydra 71.9 & 72.3 \\
InsectSound & 1\% & 62.0 & 59.0 & 42.8 & 6/6 & Quant 63.4 & 56.8 \\
 & 10\% & 71.7 & 68.0 & 56.8 & 2/2 & Hydra 75.6 & 69.2 \\
\bottomrule
\end{tabular}
\end{table}

\paragraph{Random forest instead of ridge.}
Table~\ref{tab:rf} applies Reverso's published recipe (max pooling, patch
statistics and an embedding of the differenced series) with ridge and with a
300-tree random forest \citep{auer2025frozen,reverso2026} on the pilot
datasets.

\begin{table}[!h]
\centering\footnotesize
\caption{Reverso's published recipe with ridge or a random forest (RF),
accuracy \%, folds 0--1, mean of 5 seeds, against the best baseline on the
same folds.}
\label{tab:rf}
\begin{tabular}{llrrr}
\toprule
Dataset & Labels & Ridge & RF & Best baseline \\
\midrule
UCIActivity & 1\% & 70.1 & 81.0 & 93.1 \\
 & 10\% & 90.4 & 89.1 & 97.2 \\
 & 100\% & 94.6 & 92.1 & 97.4 \\
\midrule
CrowdSourced & 1\% & 65.4 & 65.9 & 65.4 \\
 & 10\% & 69.7 & 65.6 & 72.5 \\
 & 100\% & 70.4 & 64.5 & 75.3 \\
\midrule
InsectSound & 1\% & 62.3 & 54.6 & 63.7 \\
 & 10\% & 72.0 & 65.8 & 75.6 \\
\bottomrule
\end{tabular}
\end{table}

\paragraph{Forecast-error channel.}
Table~\ref{tab:fe} gives the extra-channel runs behind Section~\ref{sec:results}.

\begin{table}[!h]
\centering\scriptsize
\caption{MiniRocket with one extra channel per input channel (accuracy \%,
folds 0--1, mean of 2 folds). FE $-$ diff.: forecast-error channel minus
first-difference channel, with the number of folds in which it is lower.}
\label{tab:fe}
\begin{tabular}{llrrrrr}
\toprule
Dataset & Labels & MiniRocket & + first diff. & + AR(5) & + forecast error & FE $-$ diff. \\
\midrule
UCIActivity & 1\% & 93.1 & 93.3 & 92.8 & 91.3 & $-2.0$ (2/2) \\
 & 10\% & 97.2 & 97.5 & 97.3 & 97.1 & $-0.4$ (2/2) \\
 & 100\% & 97.3 & 97.6 & 97.4 & 97.5 & $-0.1$ (1/2) \\
\midrule
WISDM2 & 1\% & 74.8 & 74.7 & 74.1 & 74.2 & $-0.5$ (2/2) \\
 & 10\% & 74.5 & 75.5 & 75.5 & 75.4 & $-0.1$ (1/2) \\
 & 100\% & 73.9 & 74.9 & 73.9 & 73.6 & $-1.3$ (1/2) \\
\midrule
CrowdSourced & 1\% & 59.1 & 63.5 & 62.8 & 61.5 & $-2.0$ (2/2) \\
 & 10\% & 70.4 & 71.0 & 70.9 & 69.5 & $-1.5$ (1/2) \\
 & 100\% & 72.7 & 76.6 & 76.0 & 71.9 & $-4.7$ (2/2) \\
\midrule
InsectSound & 1\% & 59.7 & 62.3 & 61.4 & 60.8 & $-1.4$ (2/2) \\
 & 10\% & 73.6 & 75.0 & 74.3 & 73.1 & $-1.8$ (2/2) \\
 & 100\% & 79.7 & 80.8 & 80.3 & 79.0 & $-1.8$ (2/2) \\
\bottomrule
\end{tabular}
\end{table}

\FloatBarrier
\paragraph{Full accuracy tables.}
Tables~\ref{tab:full1}--\ref{tab:full100} give every method at every
budget. Three patterns are not discussed in the main text. MiniRocket alone is
weak on LenDB (73.9 to 78.4\%), and one extra input channel lifts it to 84.0
to 88.1\% with a first difference and 88.0 to 92.8\% with an AR(5) residual.
Amplitude statistics alone reach 78.7 to 81.5\% on FordChallenge, within 2
points of MiniRocket at 1\%. With max pooling the pretraining effect can
reverse. On CrowdSourced at 1\%, Chronos-Bolt's random-weight copy reaches
68.0\% against 66.6\% for the pretrained model.

\begin{table}[!h]
\centering\scriptsize
\caption{Accuracy (\%) at 1\% of the training labels. Mean over folds $\pm$ half-width of the 95\% t-interval (LenDB: mean of 2 folds, no interval). $^{\ast}$ fewer folds than the other methods. -- not run. Bold: best baseline and best pretrained embedding in each column.}
\label{tab:full1}
\setlength{\tabcolsep}{3pt}
\begin{tabular}{lcccccc}
\toprule
Method & UCIAct. & WISDM2 & Crowd. & Insect. & Ford. & LenDB \\
\midrule
\multicolumn{7}{l}{\textit{Baselines}} \\
MiniRocket & \textbf{92.2} $\pm$ 2.4 & \textbf{75.3} $\pm$ 7.7 & 62.4 $\pm$ 8.9 & 59.9 $\pm$ 0.5 & 80.6 $\pm$ 1.7 & 73.9 \\
MultiRocket & 88.4 $\pm$ 0.9 & 74.9 $\pm$ 8.8 & 65.8 $\pm$ 9.8 & 62.3 $\pm$ 0.9 & 79.4 $\pm$ 1.8 & 87.6 \\
Hydra & 91.0 $\pm$ 3.3 & 72.0 $\pm$ 10.7 & 66.2 $\pm$ 10.3 & 62.5 $\pm$ 0.2 & 69.0 $\pm$ 2.7 & 85.5 \\
Quant & 84.0 $\pm$ 1.9 & 74.3 $\pm$ 8.2 & \textbf{66.3} $\pm$ 10.5 & \textbf{63.5} $\pm$ 0.7 & \textbf{84.8} $\pm$ 2.6 & \textbf{91.1} \\
\midrule
\multicolumn{7}{l}{\textit{MiniRocket with a cheap extra input, and amplitude statistics alone}} \\
MiniRocket + first diff. & 92.0 $\pm$ 2.2 & 75.6 $\pm$ 8.8 & 63.1 $\pm$ 10.9 & 62.3 $\pm$ 0.5 & 79.1 $\pm$ 1.7 & 84.0 \\
MiniRocket + AR(5) residual & 91.5 $\pm$ 1.9 & 75.4 $\pm$ 8.9 & 63.3 $\pm$ 9.1 & 61.5 $\pm$ 0.7 & 79.4 $\pm$ 1.7 & 88.0 \\
Amplitude statistics only & 66.9 $\pm$ 3.1 & 64.4 $\pm$ 4.5 & 59.6 $\pm$ 7.2 & 21.0 $\pm$ 0.6 & 78.7 $\pm$ 2.1 & 51.1 \\
\midrule
\multicolumn{7}{l}{\textit{Pretrained embeddings}} \\
Reverso, mean pool & 61.5 $\pm$ 2.0 & 67.0 $\pm$ 9.1 & 64.1 $\pm$ 5.4 & 57.4 $\pm$ 0.8 & 71.2 $\pm$ 2.7 & 86.2 \\
Reverso, max pool & 65.8 $\pm$ 2.5 & 67.3 $\pm$ 9.1 & 65.7 $\pm$ 6.5 & \textbf{59.6} $\pm$ 0.6 & 71.3 $\pm$ 3.0 & 88.3 \\
Reverso, mean pool + S & 68.3 $\pm$ 2.2 & \textbf{70.6} $\pm$ 7.8 & 64.4 $\pm$ 5.5 & 57.0 $\pm$ 0.6 & 72.4 $\pm$ 2.2 & 86.2 \\
Chronos-Bolt, mean pool & 67.8 $\pm$ 2.0 & 65.3 $\pm$ 7.6 & 65.3 $\pm$ 7.1 & 54.1 $\pm$ 1.0 & 71.8 $\pm$ 2.8 & \textbf{90.3} \\
Chronos-Bolt, max pool & 71.5 $\pm$ 1.8 & 66.4 $\pm$ 7.7 & \textbf{66.6} $\pm$ 8.1 & 57.9 $\pm$ 0.7 & 71.9 $\pm$ 2.9 & 89.3 \\
Chronos-Bolt, mean pool + S & \textbf{75.3} $\pm$ 1.7 & 69.1 $\pm$ 6.9 & 65.5 $\pm$ 6.9 & 54.0 $\pm$ 1.1 & \textbf{73.7} $\pm$ 2.6 & 90.2 \\
\midrule
\multicolumn{7}{l}{\textit{Random-weight copies}} \\
Reverso, random weights, mean & 51.1 $\pm$ 2.0 & 60.2 $\pm$ 10.8 & 61.1 $\pm$ 4.2 & 55.3 $\pm$ 0.6 & 70.0 $\pm$ 3.4 & -- \\
Reverso, random weights, max & 51.6 $\pm$ 5.4$^{\ast}$ & 61.1 $\pm$ 10.0 & 60.9 $\pm$ 9.0$^{\ast}$ & 56.7 $\pm$ 1.4$^{\ast}$ & 69.3 $\pm$ 3.5 & 77.8 \\
Chronos-Bolt, random weights, mean & 44.8 $\pm$ 1.2 & 58.5 $\pm$ 12.5 & 61.4 $\pm$ 2.9 & 49.3 $\pm$ 0.6 & 68.7 $\pm$ 3.3 & 73.5 \\
Chronos-Bolt, random weights, max & 59.2 $\pm$ 2.0 & 58.7 $\pm$ 13.6 & 68.0 $\pm$ 5.5 & 53.5 $\pm$ 0.4 & 69.7 $\pm$ 3.6 & 76.6 \\
\bottomrule
\end{tabular}
\end{table}

\begin{table}[!h]
\centering\scriptsize
\caption{Accuracy (\%) at 10\% of the training labels. Mean over folds $\pm$ half-width of the 95\% t-interval (LenDB: mean of 2 folds, no interval). $^{\ast}$ fewer folds than the other methods. -- not run. Bold: best baseline and best pretrained embedding in each column.}
\label{tab:full10}
\setlength{\tabcolsep}{3pt}
\begin{tabular}{lcccccc}
\toprule
Method & UCIAct. & WISDM2 & Crowd. & Insect. & Ford. & LenDB \\
\midrule
\multicolumn{7}{l}{\textit{Baselines}} \\
MiniRocket & \textbf{96.2} $\pm$ 3.3 & 75.5 $\pm$ 8.5 & 67.0 $\pm$ 12.3 & 73.6 $\pm$ 0.6 & 86.7 $\pm$ 1.7 & 76.4 \\
MultiRocket & 95.4 $\pm$ 3.9 & \textbf{76.5} $\pm$ 7.7 & 67.9 $\pm$ 15.4 & 74.0 $\pm$ 0.6 & 85.7 $\pm$ 1.3 & 90.1 \\
Hydra & 94.8 $\pm$ 4.4 & 71.9 $\pm$ 11.3 & \textbf{70.4} $\pm$ 10.9 & \textbf{75.5} $\pm$ 0.6 & 76.4 $\pm$ 1.5 & 88.0 \\
Quant & 91.2 $\pm$ 4.1 & 74.1 $\pm$ 8.2 & 65.7 $\pm$ 13.3 & 72.7 $\pm$ 0.5 & \textbf{90.1} $\pm$ 2.2 & \textbf{93.1} \\
\midrule
\multicolumn{7}{l}{\textit{MiniRocket with a cheap extra input, and amplitude statistics alone}} \\
MiniRocket + first diff. & 96.3 $\pm$ 3.3 & 76.3 $\pm$ 8.6 & 67.2 $\pm$ 14.8 & 74.9 $\pm$ 0.6 & 85.2 $\pm$ 1.5 & 86.9 \\
MiniRocket + AR(5) residual & 96.3 $\pm$ 3.0 & 76.7 $\pm$ 8.9 & 67.1 $\pm$ 13.7 & 74.2 $\pm$ 0.7 & 85.2 $\pm$ 1.3 & 91.3 \\
Amplitude statistics only & 75.9 $\pm$ 3.8 & 66.0 $\pm$ 5.2 & 58.8 $\pm$ 11.8 & 25.8 $\pm$ 0.7 & 80.9 $\pm$ 1.8 & 49.7 \\
\midrule
\multicolumn{7}{l}{\textit{Pretrained embeddings}} \\
Reverso, mean pool & 81.3 $\pm$ 2.0 & 70.5 $\pm$ 7.6 & 68.5 $\pm$ 9.4 & 68.8 $\pm$ 0.7 & 77.2 $\pm$ 1.8 & 89.8 \\
Reverso, max pool & 81.4 $\pm$ 2.4 & 70.3 $\pm$ 7.8 & 68.3 $\pm$ 9.3 & \textbf{69.6} $\pm$ 0.8 & 77.3 $\pm$ 1.9 & 89.9 \\
Reverso, mean pool + S & 90.7 $\pm$ 1.8 & \textbf{72.4} $\pm$ 6.8 & 68.8 $\pm$ 9.5 & 68.9 $\pm$ 0.7 & \textbf{81.3} $\pm$ 1.5 & 89.2 \\
Chronos-Bolt, mean pool & 82.8 $\pm$ 2.2 & 69.4 $\pm$ 7.0 & 67.4 $\pm$ 9.7 & 67.4 $\pm$ 0.7 & 76.7 $\pm$ 2.0 & \textbf{91.8} \\
Chronos-Bolt, max pool & 83.6 $\pm$ 2.2 & 69.6 $\pm$ 7.2 & \textbf{69.1} $\pm$ 10.1 & 66.4 $\pm$ 0.4 & 76.9 $\pm$ 1.9 & 90.9 \\
Chronos-Bolt, mean pool + S & \textbf{91.2} $\pm$ 2.0 & 71.9 $\pm$ 6.5 & 68.1 $\pm$ 9.7 & 67.3 $\pm$ 0.9 & 81.0 $\pm$ 1.6 & 91.7 \\
\midrule
\multicolumn{7}{l}{\textit{Random-weight copies}} \\
Reverso, random weights, mean & 73.8 $\pm$ 2.1 & 64.8 $\pm$ 8.9 & 65.4 $\pm$ 9.3 & 64.2 $\pm$ 0.8 & 74.9 $\pm$ 1.7 & -- \\
Reverso, random weights, max & 74.8 $\pm$ 4.9$^{\ast}$ & 65.0 $\pm$ 9.0 & 65.0 $\pm$ 26.4$^{\ast}$ & 65.0 $\pm$ 0.8$^{\ast}$ & 74.7 $\pm$ 1.9 & 79.5 \\
Chronos-Bolt, random weights, mean & 74.4 $\pm$ 1.9 & 65.2 $\pm$ 10.3 & 65.5 $\pm$ 5.8 & 61.3 $\pm$ 0.5 & 72.2 $\pm$ 2.9 & 76.6 \\
Chronos-Bolt, random weights, max & 76.5 $\pm$ 2.2 & 63.7 $\pm$ 11.2 & 68.1 $\pm$ 8.5 & 61.8 $\pm$ 0.3 & 72.8 $\pm$ 2.6 & 80.0 \\
\bottomrule
\end{tabular}
\end{table}

\begin{table}[!h]
\centering\scriptsize
\caption{Accuracy (\%) at 100\% of the training labels. Mean over folds $\pm$ half-width of the 95\% t-interval (LenDB: mean of 2 folds, no interval). $^{\ast}$ fewer folds than the other methods. -- not run. Bold: best baseline and best pretrained embedding in each column.}
\label{tab:full100}
\setlength{\tabcolsep}{3pt}
\begin{tabular}{lcccccc}
\toprule
Method & UCIAct. & WISDM2 & Crowd. & Insect. & Ford. & LenDB \\
\midrule
\multicolumn{7}{l}{\textit{Baselines}} \\
MiniRocket & 96.5 $\pm$ 2.9 & 75.0 $\pm$ 8.9 & \textbf{70.3} $\pm$ 10.2 & 79.7 $\pm$ 0.7 & 88.5 $\pm$ 1.9 & 78.4 \\
MultiRocket & \textbf{96.6} $\pm$ 3.1 & \textbf{75.8} $\pm$ 8.5 & 69.1 $\pm$ 15.2 & 80.7 $\pm$ 0.3 & 88.2 $\pm$ 1.9 & 91.4 \\
Hydra & 95.5 $\pm$ 4.0 & 71.0 $\pm$ 10.5 & 70.1 $\pm$ 13.7 & \textbf{80.8} $\pm$ 0.7 & 80.0 $\pm$ 1.2 & 88.7 \\
Quant & 91.8 $\pm$ 6.5 & 73.6 $\pm$ 8.7 & 65.5 $\pm$ 12.7 & 77.6 $\pm$ 0.3 & \textbf{93.2} $\pm$ 1.5 & \textbf{93.8} \\
\midrule
\multicolumn{7}{l}{\textit{MiniRocket with a cheap extra input, and amplitude statistics alone}} \\
MiniRocket + first diff. & 97.1 $\pm$ 2.2 & 75.5 $\pm$ 9.5 & 70.6 $\pm$ 14.6 & 80.8 $\pm$ 0.5 & 87.6 $\pm$ 1.7 & 88.1 \\
MiniRocket + AR(5) residual & 96.8 $\pm$ 2.4 & 75.5 $\pm$ 8.3 & 70.9 $\pm$ 14.8 & 80.3 $\pm$ 0.6 & 87.7 $\pm$ 1.7 & 92.8 \\
Amplitude statistics only & 82.9 $\pm$ 3.8 & 66.2 $\pm$ 5.2 & 59.2 $\pm$ 8.4 & 26.4 $\pm$ 0.8 & 81.5 $\pm$ 1.9 & 49.7 \\
\midrule
\multicolumn{7}{l}{\textit{Pretrained embeddings}} \\
Reverso, mean pool & 87.3 $\pm$ 1.3 & 71.6 $\pm$ 7.6 & 68.6 $\pm$ 7.2 & 73.4 $\pm$ 0.6 & -- & 91.1 \\
Reverso, max pool & 86.2 $\pm$ 1.5 & 71.1 $\pm$ 7.9 & 67.8 $\pm$ 9.1 & 73.2 $\pm$ 0.6 & -- & 91.2 \\
Reverso, mean pool + S & 94.6 $\pm$ 1.7 & \textbf{73.4} $\pm$ 7.3 & \textbf{70.0} $\pm$ 9.2 & \textbf{73.6} $\pm$ 0.6 & -- & 89.8 \\
Chronos-Bolt, mean pool & 88.3 $\pm$ 1.9 & 70.8 $\pm$ 7.4 & 66.7 $\pm$ 8.2 & 72.0 $\pm$ 0.9 & -- & \textbf{92.7} \\
Chronos-Bolt, max pool & 87.9 $\pm$ 1.8 & 70.5 $\pm$ 7.1 & 68.6 $\pm$ 9.0 & 70.0 $\pm$ 0.2 & -- & 91.6 \\
Chronos-Bolt, mean pool + S & \textbf{94.7} $\pm$ 2.4 & 73.3 $\pm$ 7.0 & 68.3 $\pm$ 9.8 & 72.3 $\pm$ 0.9 & -- & 92.2 \\
\midrule
\multicolumn{7}{l}{\textit{Random-weight copies}} \\
Reverso, random weights, mean & 81.3 $\pm$ 1.6 & 66.8 $\pm$ 8.4 & 66.9 $\pm$ 9.8 & 68.9 $\pm$ 1.2$^{\ast}$ & -- & -- \\
Reverso, random weights, max & 82.7 $\pm$ 3.1$^{\ast}$ & 66.7 $\pm$ 8.4 & 66.3 $\pm$ 26.8$^{\ast}$ & 68.5 $\pm$ 1.3$^{\ast}$ & -- & 80.8 \\
Chronos-Bolt, random weights, mean & 81.3 $\pm$ 1.5 & 67.4 $\pm$ 9.5 & 67.2 $\pm$ 7.1 & 66.0 $\pm$ 0.6 & -- & 77.5 \\
Chronos-Bolt, random weights, max & 81.3 $\pm$ 2.1 & 65.7 $\pm$ 10.0 & 66.9 $\pm$ 7.9 & 65.9 $\pm$ 0.5 & -- & 82.0 \\
\bottomrule
\end{tabular}
\end{table}

\FloatBarrier

\paragraph{Compute cost.}\label{apd:cost}
Table~\ref{tab:cost} gives feature extraction times. Chronos-Bolt is faster
than Reverso on every dataset despite having about 80 times more parameters,
likely because we run Reverso's recurrent layers in plain PyTorch
(Appendix~\ref{apd:impl}). The forecast-error channel is the most expensive
input, at 41 minutes to 2.7 hours per fold.

\begin{table}[!h]
\centering\scriptsize
\caption{Feature extraction time in seconds for one fold and all series
(classifier fitting excluded). Baselines ran on CPU and the forecasters on an
8\,GB laptop GPU. The forecast-error channel includes Reverso's forecasts for
every series. Fine-tuning: one training run at 10\% labels. Times were
measured while other jobs shared the machine, so they are indicative only.}
\label{tab:cost}
\begin{tabular}{lrrrrrr}
\toprule
Method & UCIAct. & WISDM2 & Crowd. & Insect. & Ford. & LenDB \\
\midrule
MiniRocket & 2 & 59 & 4 & 8 & 5 & 193 \\
MultiRocket & 4 & 51 & 8 & 26 & 9 & 283 \\
Hydra & 13 & 206 & 34 & 356 & 30 & 2{,}242 \\
Quant & 5 & 53 & 13 & 26 & 22 & 552 \\
Reverso embeddings & 353 & 1{,}904 & 656 & 193 & 4{,}848 & 28{,}033 \\
Chronos-Bolt embeddings & 31 & 140 & 114 & 70 & 183 & 2{,}458 \\
Forecast-error channel & 2{,}468 & 5{,}430 & 9{,}840 & 7{,}103 & -- & -- \\
Fine-tuning Reverso & 3{,}389 & -- & 5{,}824 & 2{,}324 & -- & -- \\
\bottomrule
\end{tabular}
\end{table}

\end{document}